\documentclass[letterpaper, 10 pt, conference]{ieeeconf}  
\usepackage[automake]{glossaries-extra}
\usepackage[maxbibnames=99, isbn=false, doi=false, eprint=false]{biblatex}
\AtEveryBibitem{
  \clearfield{month}
  \clearfield{urlyear}
  \clearfield{urlmonth}
  \clearfield{urlday}
  \ifboolexpr{test {\ifentrytype{inproceedings}} or test {\ifentrytype{incollection}}}{
    \clearlist{publisher}
    \clearlist{location}
  }{}
}
\usepackage{float}
\usepackage{subcaption}
\usepackage{lipsum}
\usepackage{placeins}
\usepackage{dblfloatfix}

\usepackage{xcolor}
\definecolor{boxblue}{HTML}{DAE8FC}
\definecolor{boxpurple}{HTML}{E1D5E7}
\definecolor{boxyellow}{HTML}{FFF2CC}
\definecolor{boxred}{HTML}{F8CECC}
\definecolor{boxgreen}{HTML}{D5E8D4}
\definecolor{boxgray}{HTML}{E6E6E6}

\usepackage[most]{tcolorbox}
\newtcbox{\callbox}[1]{on line, boxrule=0pt, boxsep=0pt, top=0pt, bottom=-1pt, left=3pt, right=3pt, colback=#1, arc=0pt, outer arc=0pt, before upper=\strut}
\newtcbox{\valbox}[1]{on line, boxrule=0pt, boxsep=0pt, top=0pt, bottom=0pt, left=3pt, right=3pt, colback=#1, arc=1ex, auto outer arc, before upper=\strut}

\usepackage{censor}

\usepackage[frozencache,cachedir=.]{minted}
\setminted{fontsize=\small}

\usepackage{xspace}

\usepackage{graphics} 
\usepackage{epsfig} 
\usepackage{mathptmx} 
\usepackage{times} 
\usepackage{amsmath} 
\usepackage{amssymb}  
\usepackage{hyperref}

\setabbreviationstyle[acronym]{long-postshort-user}
\glssetcategoryattribute{acronym}{nohyperfirst}{true}
\setabbreviationstyle{short-nolong}

\newglossaryentry{asyncio}{name=Asyncio, description={A Python library for asynchronous code.}}
\newglossaryentry{cuda}{name=CUDA, description={A Python library for asynchronous code.}}
\newglossaryentry{symengine}{name=SymEngine, description={A Python library for asynchronous code.}}
\newglossaryentry{sympy}{name=SymPy, description={A Python library for asynchronous code.}}
\newglossaryentry{symforce}{name=SymForce, description={}}
\newglossaryentry{cudss}{name=CuDSS, description={A Cuda framework for solving large sparse systems of linear equations.}}

\newacronym{asv}{ASV}{Autonomous Surface Vehicle}
\newacronym{dolp}{DoLP}{Degree of Linear Polarization}
\newacronym{aolp}{AoLP}{Angle of Linear Polarization}
\newacronym{sitaw}{SITAW}{Situational Awareness}
\newacronym{cse}{CSE}{Common Subexpression Elimination}
\newacronym{cpse}{CSE}{Common Partial Subexpression Elimination}
\newacronym{fma}{FMA}{Fused Multiply Add}
\newacronym{poe}{PoE}{Power over Ethernet}
\newacronym{pps}{PPS}{Pulse Per Second}
\newacronym{pcgnr}{PCGNR}{Preconditioned Conjugate Gradient with Normal Residual}
\newacronym{bal}{BAL}{Bundle Adjustment in the Large}
\newacronym{mpc}{MPC}{Model Predictive Control}
\newacronym{mojo}{Mojo}{A new programming language}
\newacronym{dabseg}{DABSEG}{Directed Acyclic Bipartite Symbolic Expression Graph}
\newacronym{mse}{MSE}{Mean Squared Error}
\newacronym{pcg}{PCG}{Preconditioned Conjugate Gradient}
\newacronym{sass}{SASS}{Source and Assembly}

\glsaddall
\makeglossaries

\IEEEoverridecommandlockouts                              
\title{\LARGE \bf
Caspar: CUDA Accelerator for Symbolic Programming \\with Adaptive Reordering
}

\author{Emil Martens$^{1,2}$, Aaron Miller$^{2}$, Matias Varnum$^{2}$ and Annette Stahl$^{1}$
\thanks{$^{1}$ Norwegian University of Science and Technology}%
\thanks{$^{2}$ Skydio \url{https://www.skydio.com/} }%
}

\begin{document}

\maketitle
\thispagestyle{empty}
\pagestyle{empty}


\begin{abstract}
	We present Caspar, a library that makes the power of modern GPUs more accessible in robotics and provides a state-of-the-art nonlinear GPU solver that can be applied to a wide range of different optimization problems.
	Caspar bridges the gap between expressive symbolic programming in Python and high-performance GPU runtimes in C++ by automatically generating optimized CUDA kernels from symbolic expressions.
	Building on the SymForce library, users can easily define and combine symbolic expressions, including Lie group operations, to generate custom CUDA kernels.
	To use Caspar as a solver, users need only define the symbolic residual functions; Caspar then uses symbolic differentiation to generate the necessary GPU kernels and interfaces to perform nonlinear optimization.

	In this paper, we present the core components of Caspar and showcase its performance by performing bundle adjustment on the \gls{bal} dataset.
	We benchmark Caspar against other state-of-the-art bundle adjusters and show that it is 5 to 20 times faster than the best alternative, requires less memory, and achieves similar accuracy.
	This illustrates the benefit of our symbolic GPU programming approach.
	Caspar is released as part of SymForce and is freely available at \url{https://github.com/symforce-org/symforce}.
\end{abstract}

\section{Introduction and Related Work}

Many core tasks in robotics, like motion planning, state estimation, and sensor calibration, are naturally formulated as nonlinear optimization problems \cite{boyd2004convex,nocedal2006numerical}.
While constrained nonlinear optimization requires the solution to satisfy specific equality or inequality constraints (e.g. collision avoidance or robot joint limits), unconstrained optimization seeks to minimize the cost of problems without such restrictions.
The latter is generally preferred in robotics, particularly in real-time systems, due to its relative simplicity and computational efficiency, especially when the problem-specific structure allows constraints to be handled implicitly \cite{lavalle2006planning,siciliano2016springer}.

Unconstrained nonlinear least-squares problems are often solved using the Levenberg-Marquardt algorithm \cite{madsenMETHODSNONLINEARLEAST2004}.
To solve the large, sparse linear systems that arise from these problems,  \gls{pcg} is commonly used, due to its convergence properties and low memory requirements, especially when combined with effective preconditioning \cite{saad2003iterative}.
These methods can be applied in a wide range of real-world robotics applications, such as bundle adjustment \cite{triggs2000bundle}, pose graph optimization \cite{kuemmerle2011g2o}, and simultaneous localization and mapping (SLAM)~\cite{dellaert2006square,cadena2016past}, where constraints are typically handled outside the core solver loop or implicitly encoded via robust cost functions \cite{RomeInaDay}.

Unconstrained nonlinear problems thus play a central role in robotics optimization, as they can represent complex estimation tasks while remaining computationally efficient to solve.
While existing algorithms and implementations exist to solve these types of problems, Caspar offers a new way to generate solvers from symbolic residual functions that leverage the power of the GPU to deliver state-of-the-art performance.
This paper covers the theory and design of Caspar, showcasing how it is able to generate a bundle adjuster that is \textbf{5 to 20 times faster} than relevant alternatives.


\subsection{Symbolic Computing}

In order to solve nonlinear optimization problems efficiently, we need a function to efficiently compute the least-squares residual and its Jacobian.
SymForce \cite{martirosSymForceSymbolicComputation2022} is a library that allows the user to write a symbolic definition of their problem in Python, with access to primitive operations from SymPy, as well as more complex types such as Lie groups and other manifolds.
Once the symbolic representation is defined, the library can be used to compute the necessary derivatives and generate efficient code to compute them at runtime in multiple target languages.
Having support for manifolds makes it trivial to compute derivatives of functions on manifolds and to use them to perform on-manifold optimization.
SymForce uses symbolic differentiation to compute derivatives, allowing it to leverage fine-grained sparsity in intermediate Jacobians.
Symbolic differentiation in SymForce does not suffer from exponential complexity at either code generation time or runtime \cite{laueEquivalenceAutomaticSymbolic2022, martirosSymForceSymbolicComputation2022}.
It is equivalent to automatic differentiation in that aspect.

When generating code from SymForce expressions, multiple techniques are used to produce more efficient runtime functions than are generally produced by automatic differentiation, and by typical compilers and software design practices, because SymForce flattens the entire linearization process into a single runtime function, performs common subexpression elimination on the entire function, exploits fine-grained sparsity, and allows for more complex algebraic simplifications that compilers generally cannot or will not do automatically \cite{martirosSymForceSymbolicComputation2022,rawatAssociativeInstructionReordering2018}.

\subsection{GPU Computing}

A modern GPU is capable of simultaneously executing hundreds of thousands of threads, significantly outperforming CPUs in many tasks.
However, since GPUs are fundamentally different in architecture from CPUs, it is crucial to optimize code specifically for how GPU hardware operates.
An NVIDIA GPU consists of multiple independent execution units called SMs (Streaming Multiprocessors).
As a CUDA program is launched, the individual blocks of the kernel will be distributed to the available SMs by the GPU scheduler.
The ability of an SM to accept additional blocks depends on whether it has sufficient resources to fit the new block \cite{howCudaProgrammingWorks}.
There are many different resources that are part of this consideration, but the ones most relevant to this paper are the total number of threads a SM is capable of running at once, the total number of registers available, and the total amount of shared memory.
Therefore, reducing the number of registers or shared memory needed by a CUDA kernel can have a dramatic effect on the total runtime of the kernel because each SM can fit additional blocks at once.

Additionally, with the large amount of compute available on a GPU, the total memory bandwidth can quickly become the limiting factor for the execution speed of a CUDA program \cite{howCudaProgrammingWorks}.
When main memory is read, it is always read in blocks called a page.
It is therefore highly beneficial to construct the memory access pattern of a CUDA warp such that the memory read by each thread is packed tightly together.
This allows us to fetch all the required memory for the entire warp with a single page read.
If each thread reads from memory locations that are far apart, we may, in the worst case, need to read an entire page for each thread, despite only needing a small portion of it.
This can end up achieving as little as 8\% of peak memory bandwidth \cite{howCudaProgrammingWorks}.

Memory bandwidth frequently acts as the primary bottleneck in GPU applications, making careful optimization of memory access patterns essential \cite{howCudaProgrammingWorks}.
Sharing data between threads, either through shared memory or warp-level instructions, is an effective strategy to avoid redundant memory accesses \cite{nvidiaCUDABestPractices2025}.
Another key optimization technique is kernel fusion, which combines multiple dependent kernels into a single operation, eliminating the need to read and write intermediate data between kernel invocations.

To efficiently harness the computational power of GPUs, a wide range of libraries and frameworks have been developed.
High-level frameworks, like PyTorch provide intuitive interfaces for executing tensor operations from Python and building efficient computational pipelines.
Meanwhile, task-specific libraries such as cuDSS and cuSPARSE deliver highly optimized performance within their respective domains.
Caspar contributes to this ecosystem by introducing symbolic programming as a novel interface for GPU acceleration, offering excellent performance for sparse nonlinear optimization problems.

\section{Symbolic Programming with Caspar}
\label{sec:symbolic_programming}
Caspar builds on \gls{symforce}, a library for working with symbolic Lie groups and sensor models used in robotics \cite{martirosSymForceSymbolicComputation2022}.
After defining symbolic expressions with \gls{symforce}, Caspar converts them into a \gls{dabseg}, enabling symbolic optimizations and the generation of efficient \gls{cuda} kernels.
The \gls{dabseg} is composed of \mintinline{python}{call} and \mintinline{python}{value} nodes, as illustrated in Figure~\ref{fig:codegen-6}.
The \mintinline{python}{call} nodes represent the invocation of symbolic functions and may include private data, such as literal values used in the function call.
The \mintinline{python}{value} nodes represent the outputs of the \mintinline{python}{call} nodes and are allocated to registers.
This intermediate representation provides several benefits, most notably support for symbolic functions with multiple outputs and efficient graph manipulation.
After generation, the \gls{dabseg} is analyzed, optimized, topologically sorted, and then mapped to CUDA instructions to produce high-performance CUDA kernels.
Below are some of the most important symbolic optimizations performed by Caspar.

\subsection{Hardware-Mapped Functions}
Caspar reformulates symbolic expressions to take advantage of dedicated hardware instructions and guarantees their use whenever possible.
For instance, the dedicated \mintinline{cuda}{norm[n]} and \mintinline{cuda}{rnorm[n]} instructions are used whenever the expression tree contains the norm or normalization operation of a vector of three or four elements, which is common in robotics applications.
Other nontrivial hardware-mapped functions include \mintinline{cuda}{ldexp} for $x*2^k, \; k\in\mathbb{Z}$, \mintinline{cuda}{atan2}, \mintinline{cuda}{fma} (see \ref{sec:accumulators}), \mintinline{cuda}{sincos} (see \ref{sec:context_aware_reformulation}), and the ternary operator.

Following the CUDA C++ Programming Guide \cite[68-69]{nvidiaCUDABestPractices2025}, Caspar will decompose exponentials into hardware-accelerated components when possible.
For instance, $x^{-1/6}$ is evaluated as $1/\sqrt[3]{\sqrt{x}}$ using \mintinline{cuda}{rcbrtf(sqrtf(x))}.
Caspar automatically generates a lookup table using
\mintinline{cuda}{mulf} ($x \cdot y$),
\mintinline{cuda}{rcpf} ($1/x$),
\mintinline{cuda}{sqrtf} ($\sqrt{x}$),
\mintinline{cuda}{rsqrtf} ($1/\sqrt{x}$),
\mintinline{cuda}{cbrtf} ($\sqrt[3]{x}$), and
\mintinline{cuda}{rcbrtf} ($1/\sqrt[3]{x}$),
based on the given cost of each instruction.
The regular \mintinline{cuda}{expf} ($x^y$) is used as a fallback only when efficient decomposition is not possible.

\subsection{Common Partial Subexpression Elimination}
In addition to regular \gls{cse} \cite{martirosSymForceSymbolicComputation2022}, Caspar eliminates partial subexpressions from commutative function calls such as sums and products, to reduce computational load.
This is done by repeatedly eliminating partial subexpressions that appear in multiple calls.
The partial subexpression that occurs most frequently is always eliminated first; ties are resolved by selecting the one with more elements.
Consider the following three sums:
\begin{minted}[escapeinside=||]{python}
  r0=a+b+c, r1=a+c+d+e, r2=a+c+e
\end{minted}
The subexpressions \mintinline{python}{{a+c}}, \mintinline{python}{{a+e}}, \mintinline{python}{{c+e}}, and \mintinline{python}{{a+c+e}} are considered for elimination because they all appear more than once.
As \mintinline{python}{{a+c}} appears three times, i.e. most frequently, it is picked for elimination.
A temporary value \mintinline{python}{r3} is created and used to eliminate \mintinline{python}{{a+c}} from \mintinline{python}{r0}, \mintinline{python}{r1}, and \mintinline{python}{r2}, yielding:
\begin{minted}[escapeinside=||]{python}
  r0=|\color{purple}r3|+b,  r1=|\color{purple}r3|+d+e,  r2=|\color{purple}r3|+e, |\color{purple}r3|=a+c
\end{minted}
After step one, the only common subexpression is \mintinline{python}{{r3+e}}.
Since \mintinline{python}{r2=r3+e}, there is no need for a temporary variable and \mintinline{python}{r2} is used directly to eliminate \mintinline{python}{{r3+e}} from \mintinline{python}{r1}.
\begin{minted}[escapeinside=||]{python}
  r0=r3+b,  r1=|\color{teal}r2|+d,    |\color{teal}r2|=r3+e, r3=a+c
\end{minted}
At the end, there are no subexpressions to eliminate, and the process terminates.

To validate the impact, we compiled the original and Caspar versions of the equations above using \texttt{nvcc}, with aggressive optimizations (\texttt{-O3}, \texttt{-use\_fast\_math}), and inspected the \gls{sass}.
The following is a comparison between what \textit{nvcc} was able to optimize without Caspar and what we obtained using Caspar:
\begin{minted}[escapeinside=||]{python}
original r0=a+b+c, r1=a+c+d+e, r2=a+c+e
nvcc     r0=a+b+c, r1=|\color{purple}r3|+d+e,  r2=|\color{purple}r3|+e, |\color{purple}r3|=a+c
caspar   r0=|\color{purple}r3|+b,  r1=|\color{teal}r2|+d,    |\color{teal}r2|=|\color{purple}r3|+e, |\color{purple}r3|=a+c
\end{minted}
The compiler's built-in \gls{cse} was able to eliminate one instance of \mintinline{python}{a+c} reducing the original kernel from 7 to 6 \texttt{FADD} instructions.
In contrast, the Caspar-generated kernel compiles to exactly 4 \texttt{FADD} instructions, demonstrating the benefit of common partial subexpression elimination.

\subsection{Context-Aware Optimizations}
\label{sec:context_aware_reformulation}
Caspar takes the broader symbolic context into account when performing symbolic optimizations.
This allows it to apply transformations that reduce the overall computational load, even if some individual operations appear more expensive in isolation.
By symbolic context, we mean the collection of symbolic expressions used to generate a kernel.
Consider the division operation $a/x$, which is most efficiently evaluated with the instruction \mintinline{cuda}{divf(a,x)}. If, however, the expression $b/x$ also appears in the same context, it is more efficient to compute the reciprocal using the hardware instruction \mintinline{cuda}{rcpf}, store it as $r = 1/x$, and then evaluate $r \cdot a$ and $r \cdot b$. This avoids performing two separate divisions, which are significantly more expensive than multiplications.

Since Caspar supports symbolic functions with multiple outputs, it can further optimize by fusing related expressions. For instance, if both $\sin$ and $\cos$ are required with the same argument, Caspar generates a single \mintinline{cuda}{sincosf} call instead of two separate trigonometric calls.




\begin{figure}
	\centering
	\includegraphics[trim={2.5cm 5cm 16cm 4cm},clip,width=0.5\textwidth]{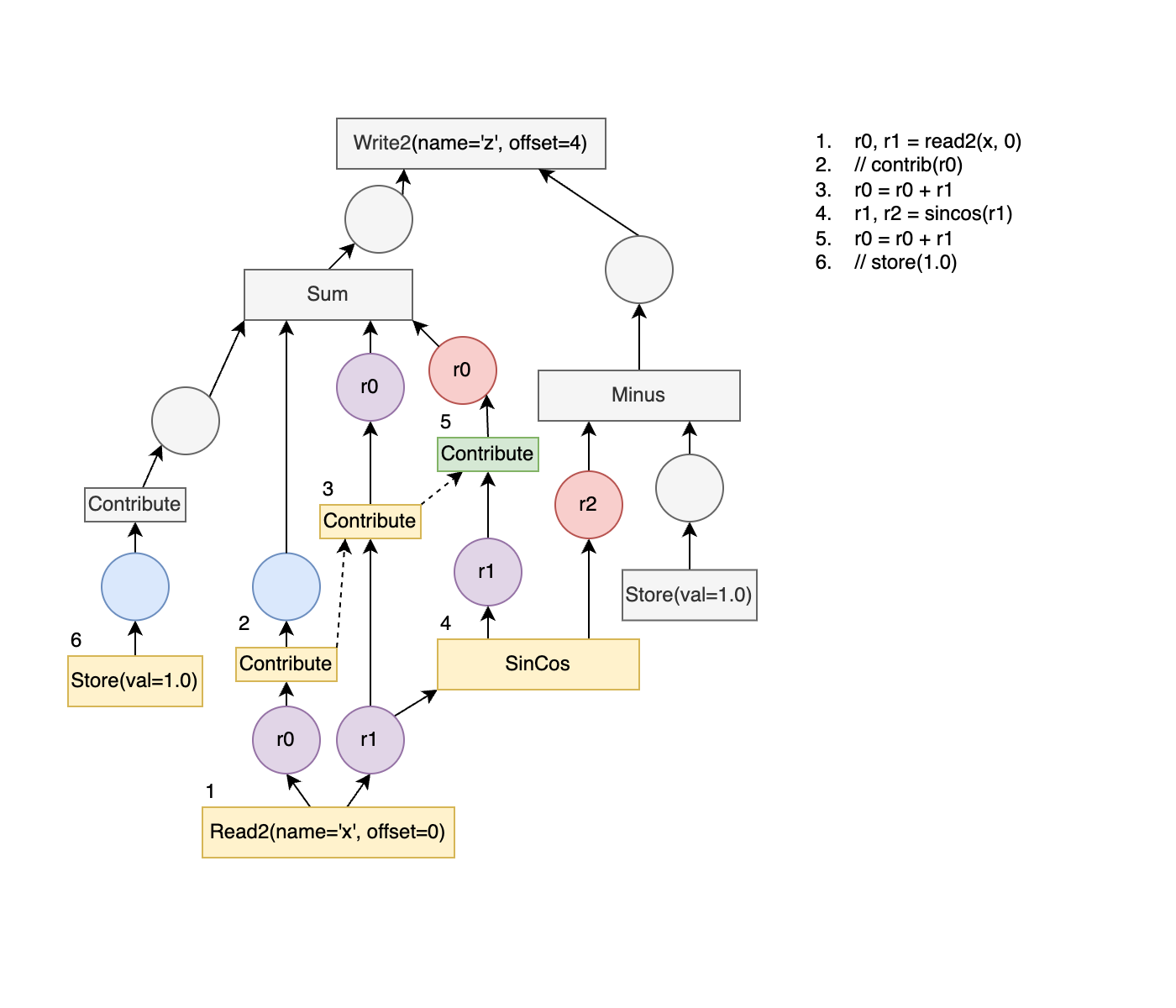}
	\caption{\glsfirst{dabseg} with partially completed topological ordering, from \mintinline{python}{[sin(x1)+x0+x1+1, cos(x1)-1]}.
		Call nodes are shown as \callbox{boxyellow}{completed}, \callbox{boxgreen}{latest}, and \callbox{boxgray}{remaining}.
		Value nodes are shown as \valbox{boxred}{live}, \valbox{boxpurple}{finished}, \valbox{boxblue}{virtual} and \valbox{boxgray}{remaining}, with assigned registers.
		The dotted arrows indicate the tracked contribution order to the \mintinline{python}{Sum} accumulator.
		\label{fig:codegen-6}}
\end{figure}

\section{Symbolic Reordering}
Caspar performs a topological ordering of the \gls{dabseg} to reduce register requirements.
The number of registers used by a kernel has a significant impact on performance \cite[141]{nvidiaCUDAProgrammingGuide2025}.
When a kernel requires more registers than are available, local memory is used instead \cite[147]{nvidiaCUDAProgrammingGuide2025}. This register spilling degrades performance because local memory is as slow as global memory \cite[147]{nvidiaCUDAProgrammingGuide2025}.

To minimize spilling, we seek an expression ordering that lowers register pressure. Although modern compilers are generally effective at controlling register usage, they tend to struggle with computational patterns that involve extensive many-to-many data reuse, the exact kind introduced by \gls{cse} when eliminating redundant computations \cite{rawatAssociativeInstructionReordering2018}.





\subsection{Ordering Heuristics}
Because register allocation and instruction scheduling are NP-hard, we use a greedy heuristic to find a good ordering of calls in the expression tree \cite{motwaniCombiningRegisterAllocation1995}.
At each step, Caspar maintains a set of firable function calls (i.e., calls whose inputs are all available).
To decide which call to schedule next, Caspar evaluates a sort key for each candidate and sorts them lexicographically.
Similar to the method proposed by Rawat et al. \cite{rawatAssociativeInstructionReordering2018}, we design this key to minimize register pressure by preferring calls that free registers or compute values that will soon be used together.
The key consists of the tuple $(R_0,A_0,R_1,A_1,R_2,A_2,R_3,A_3,K)$, where the components are defined as follows:

\subsubsection*{$R_i$} The release potential of degree $i$.
$R_0$ is the number of variables that can be freed by the call.
For the other $R_i$ values, $i$ is the number of function calls between the current call and the call that will free $R_i$ variables.

\subsubsection*{$A_i$} The affinity of degree $i$.
$A_0$ is the number of currently live variables that are used as arguments together with the output of the call in at least one call.
As with the $R_i$ values, $i$ is the number of function calls between the current call and the call that has affinity with $A_i$ variables.

\subsubsection*{${K}$} An arbitrary key value for each call to keep the ordering deterministic.
Currently, this value is the order of the call in a depth-first traversal of the expression graph, but it can be used to generate and test different orderings.

\subsection{Accumulators}
\label{sec:accumulators}
Caspar uses what we call accumulators to represent symbolic reduction operations such as sums and products, enabling their order to be optimized.
Accumulators consist of a collection of binary, commutative, associative operations, e.g., a sum is a collection of additions.
Since the binary operators can contribute independently to the parent accumulator, we do not need all the arguments of an accumulator sum to be stored in registers simultaneously.
A trivial approach to decompose the accumulators is to use a predefined order of binary operations, but this imposes artificial constraints on symbolic reordering and may prevent finding the optimal ordering.
Instead, Caspar uses \mintinline{python}{contribute} calls, where a single value is contributed to the accumulator output.
Each \mintinline{python}{contribute} call performs a binary operation, depending on the accumulator type, on the accumulator output and its input value, except for the first contribute call, which initializes the accumulator output.
The parent accumulator is only used during symbolic processing and is not translated to an instruction.
During symbolic reordering, the order of contributions is tracked, represented by the dotted arrows in Figure~\ref{fig:codegen-6}, to determine which registers to use during kernel generation.

Special logic is used to handle sums of products, ensuring that \gls{fma} operations are used whenever possible.
Consider the expression $x y + z + i j k + m$.
To utilize both possible \gls{fma} operations, either $+z$ or $+m$ must be the first contribution to the sum.
Therefore, there is no benefit in loading $x$ or $y$ before contributing either $z$ or $m$.
The $i j k$ product can be started at any time, but cannot be completed until either $+z$ or $+m$ is contributed.
Since any of these variables might be used elsewhere in the expression tree, this type of indirect interdependence significantly increases the complexity of the reordering problem.
When the sum contains only products, the problem is simpler, as the number of \gls{fma} operations does not depend on the contribution order.

\section{Symbolic Memory}
\label{sec:symbolic_memory}
Memory operations are the most critical factors for performance in CUDA, and have therefore been a primary focus during Caspar's development \cite[35]{nvidiaCUDABestPractices2025}.
Caspar incorporates all memory operations, i.e., read, write, and store, into the \gls{dabseg}.
This streamlines symbolic reformulation and enables the \gls{cse} pass to eliminate redundant read operations.
It also allows for lazy read operations and eager store operations, reducing register pressure.

\subsection{Memory Accessors}
Caspar provides symbolic memory accessors that enable efficient data access patterns tailored to different scenarios.
Each accessor type uses custom CUDA kernels, leveraging shared memory and warp instructions, to efficiently perform memory transactions between global memory and registers.
When defining symbolic kernels, users specify the appropriate access pattern for each argument and output based on the expected data layout and access pattern.
Currently, there are three families of access patterns: read, write, and add (i.e., increment). The following access patterns are supported:

\subsubsection*{Sequential Read/Write}
Each thread accesses data at its global thread index, enabling straightforward parallel processing of arrays.

\subsubsection*{Indexed Read/Write/Add}
Threads access data at user-specified indices, supporting sparse access patterns. Requires unique indices per thread to avoid race conditions when writing or adding. See Shared Add below for handling non-unique indices.

\subsubsection*{Unique Read/Add}
All threads access the same memory location, useful for broadcasting constants or accumulating global values.

\subsubsection*{Sum Write/Add}
Computes a global reduction across all threads and writes or adds the result to a single location.

\subsubsection*{Pairwise Read}
Each thread reads data from both its global index and the next consecutive location.

\subsubsection*{Pairwise Write/Add}
Threads combine their values with data from the previous thread before writing.

\subsubsection*{Shared Read/Add}
\label{sec:shared_read_add}
The shared read and add access patterns leverage shared memory and native warp instructions to improve efficiency when multiple threads within a block read or increment the same value, and are used extensively by the solver.
For shared read operations, the sorted set of unique indices is accessed collaboratively and distributed within the block, resulting in fewer memory transactions.
For shared add operations, Caspar utilizes warp instructions and shared memory to efficiently calculate the total value for each unique index within each thread block, before incrementing the corresponding global values collaboratively using atomic add. This is more efficient than using atomic add for each thread, as the access pattern of the atomic operations is sorted, the number of atomic operations is equal to the number of unique indices, and it avoids collisions between atomic operations within a thread block.
Both shared read and add rely on arrays of custom indexing structs, which are passed as arguments.
These arrays are efficiently generated from a given array of indices using a custom CUDA kernel and can be reused across multiple kernels.
Each indexing struct in the array has the following structure:
\begin{minted}{cuda}
struct SharedIndex {
  // Sorted block-unique index
  uint32_t unique;
  // Local unique index for each thread
  uint16_t target;
  // Used to minimize warp divergence
  // during partial reduction operations
  uint16_t argsort;
};
\end{minted}

\subsection{Vector Access and Blocked Struct of Arrays}
\begin{figure}[H]
	\centering
	\includegraphics[trim={12.5cm 7.6cm 1cm 0.2cm},clip,width=.48\textwidth]{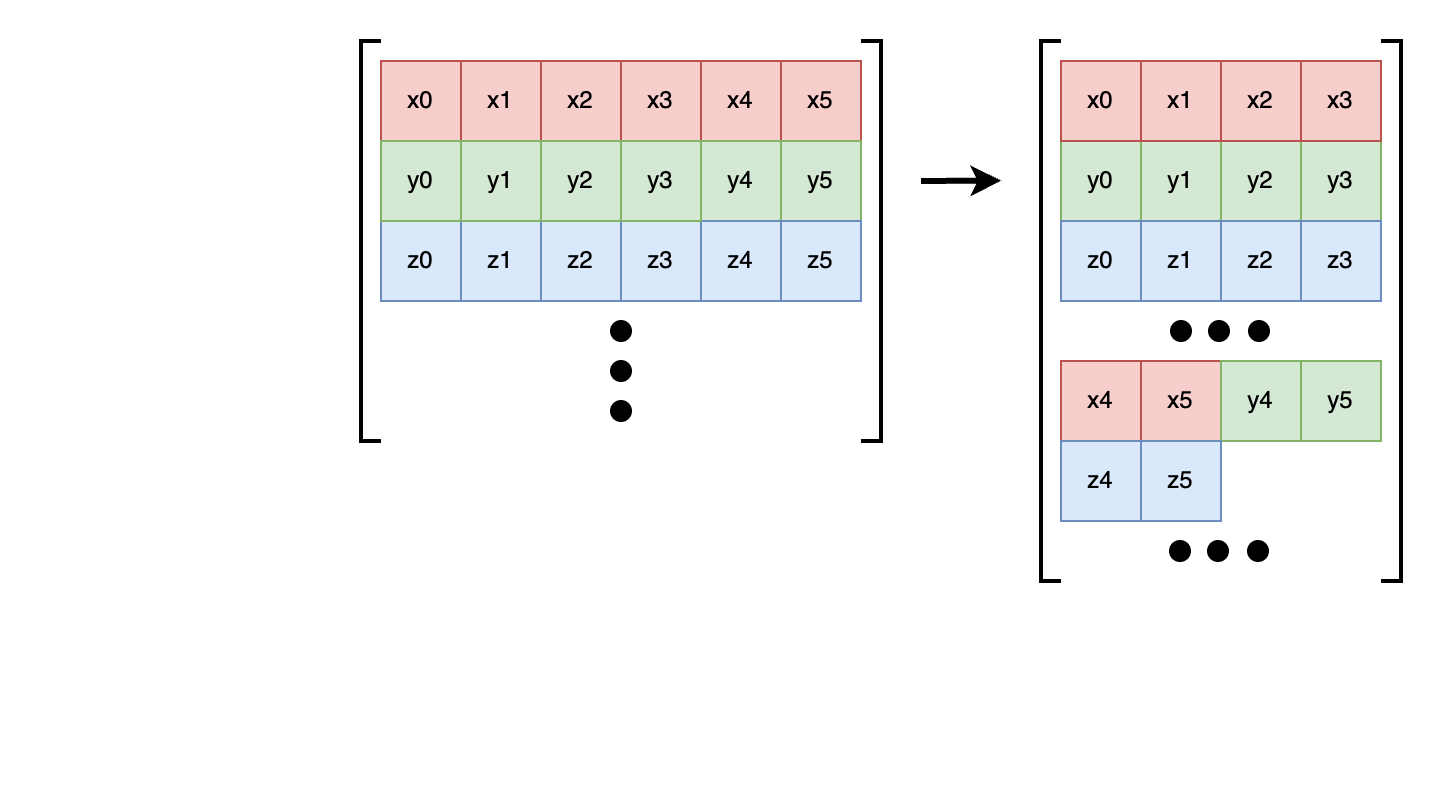}
	\caption{Mapping of an array of structs of size 6 to the corresponding blocked struct of arrays. \label{fig:memlayout-6}}
\end{figure}

As Caspar supports symbolic functions with multiple inputs and outputs, it provides symbolic memory operations that map to vector instructions in CUDA.
To maximize the utilization of vector instructions, Caspar also offers a blocked struct of arrays memory layout, illustrated in Figure~\ref{fig:memlayout-6}.
In this layout, data is stored in chunks of four elements, with any remaining values appended at the end as chunks of one, two, or three elements.
If the remaining chunk has three elements, it is padded to align to 16 bytes, since CUDA does not support vector instructions for 12-byte-aligned or 12-byte-sized elements \cite[146, 167]{nvidiaCUDAProgrammingGuide2025}.
Each block is then accessed using vector instructions of the corresponding size.
By default, Caspar uses the blocked struct of arrays layout for all data types, ensuring maximal use of vector instructions.
Caspar automatically generates kernels to map between an array of structs and a blocked struct of arrays for all relevant object types, removing the need for manual intervention.
This generation builds on the \textit{Storage} framework of \gls{symforce}, which defines how all symbolic objects are mapped to and from memory \cite{martirosSymForceSymbolicComputation2022}.

\section{Solver Design}
Caspar uses the Levenberg-Marquardt algorithm with a quality-based dampening update, as described by Madsen et al. \cite{madsenMETHODSNONLINEARLEAST2004}.
Linearization and retraction of update steps follow the tangent space differentiation approach from SymForce \cite{martirosSymForceSymbolicComputation2022}.
Following the findings of Wu et al. \cite{wuMulticoreBundleAdjustment2011}, Caspar does not rely on Schur decomposition and instead solves the linear normal equations directly.
The linear problem is preconditioned with a block Jacobi preconditioner, using one block per node in the problem, such as a pose or landmark.


\subsection{Performance Optimizations}
Caspar generates a specialized solver for each problem, defined by their collection of factor types, such as an odometry factor and a reprojection factor.
These factors are formulated from residual functions, e.g., odometry error and reprojection error, taking tunable node values, e.g., camera poses and landmark positions, as arguments.
While the factor and node argument types are fixed at generation time, the number of instances is dynamic at runtime.
The structural information available during generation is utilized to generate custom CUDA kernels for every operation used by the solver,
and kernel fusion is performed where possible to minimize memory traffic and maximize occupancy.
The generated kernels benefit from Caspar's symbolic optimizations and are particularly efficient at dealing with sparse or partially linear Jacobian blocks.
The generated solver allocates a single block of GPU memory for the entire optimization, and all utilized arrays are views into this block.
This approach allows large contiguous chunks of memory to be set to zero or copied efficiently, and several sections of the memory block are reused to minimize the memory footprint.
Separate kernels are used for the first iteration of the solver, as well as the first iteration of each \gls{pcgnr} loop, to maximize kernel fusion.
For example, in the first iteration, the initial score and the Jacobian are calculated simultaneously, while subsequent runs use the score values from the previous iteration.


\begin{figure*}[b]
	\centering
	\includegraphics[width=\textwidth]{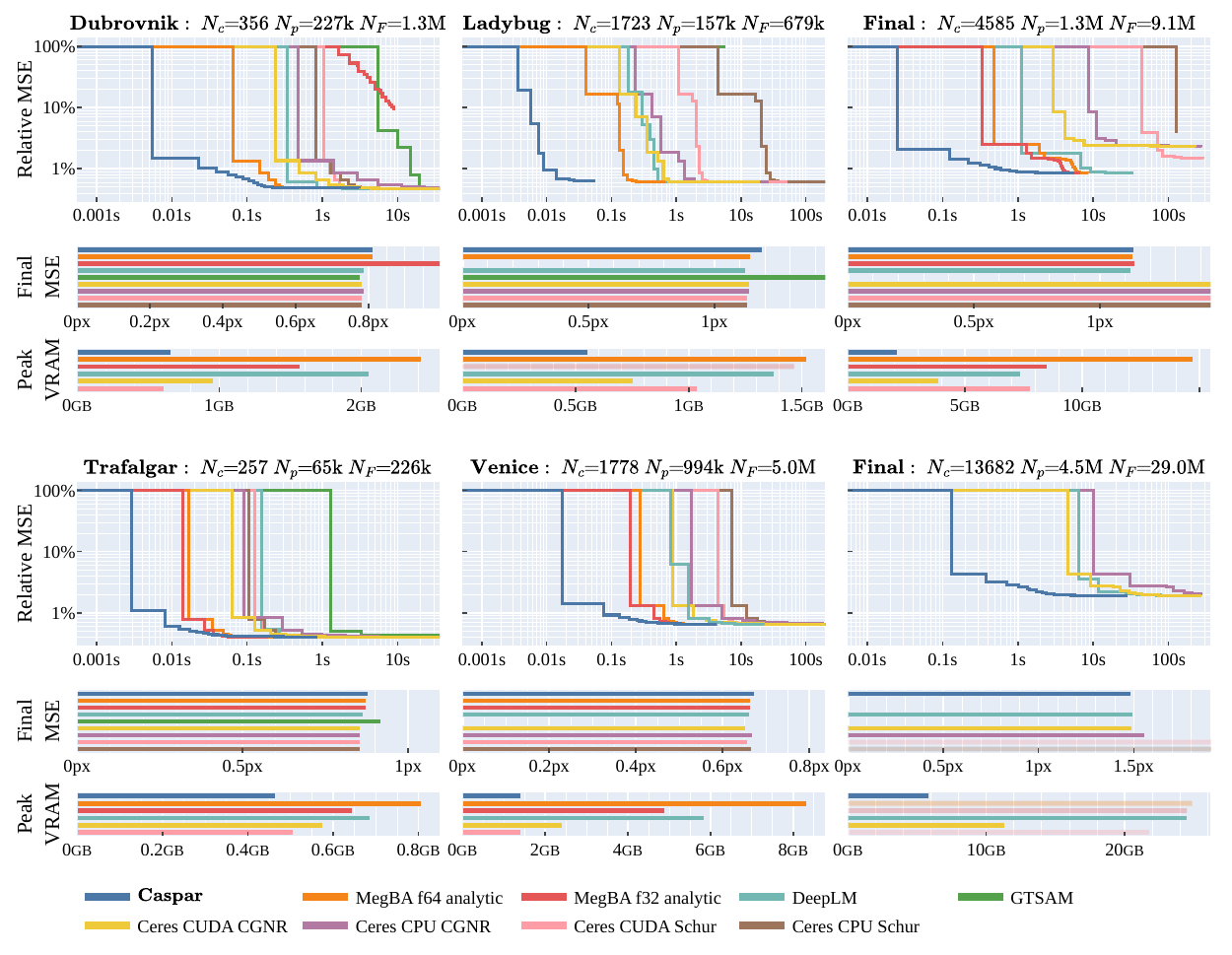}
	\caption{
		Relative \gls{mse} over time, GPU memory usage and final \gls{mse} for different solvers on multiple \gls{bal} datasets, characterized by $N_c$ cameras, $N_p$ points, and $N_F$ reprojection factors.
		Solver parameters such as initial trust region radius and inner iterations are set identically where applicable.
		Solvers that exhaust memory or fail to complete a single iteration within 300 seconds are omitted. Note that certain final MSE values go outside the chart.
		\label{fig:performance}}
\end{figure*}
\section{Experimental Results}
To demonstrate the utility and performance of Caspar, we use it to generate a bundle adjuster and perform bundle adjustment on the \gls{bal} dataset.
We compare the performance of Caspar against variations of Ceres, DeepLM, and MegBA on six different datasets from the \gls{bal} benchmark
\cite{wuMulticoreBundleAdjustment2011, huangDeepLMLargescaleNonlinear2021,renMegBAGPUBasedDistributed2022, hutchisonBundleAdjustmentLarge2010}.
We used the available example implementations of these solvers, and set the initial trust region and conjugate gradient tolerance to $100$ and $0.001$, respectively, for Caspar, Ceres, and MegBA, which yielded better results than the default values on the larger datasets.
We also considered comparing against DABA, but as it is focused on multi-GPU performance, it was not included in the comparison \cite{fanDecentralizationAccelerationEnables2023}.

The evaluation was performed on a workstation with an Intel i9-13900KF CPU, 64GB of RAM, and a NVIDIA RTX 4090 with 24GB of VRAM.

\subsection{Generating a Bundle Adjuster with Caspar}
The \gls{bal} problem contains a single type of factor, the Snavely reprojection factor, which can easily be implemented from Python, using Caspar and SymForce, as follows:
\begin{minted}[fontsize=\fontsize{9pt}{10pt}\selectfont]{python}
@dataclass
class Cam:
   cam_T_world: sf.Pose3
   calibration: sf.V3
class Point(sf.V3): ...
class Pixel(sf.V2): ...

@caslib.add_factor
def snavely_reprojection_residual(
   cam: T.Annotated[Cam, TunableShared],
   point: T.Annotated[Point, TunableShared],
   px_m: T.Annotated[Pixel, ConstSequential],
) -> sf.V2:
   focal_length, k1, k2 = cam.calibration
   point_cam = cam.cam_T_world * point
   d = point_cam[2]
   d_safe=d+sf.epsilon()*sf.sign_no_zero(d)
   p = -sf.V2(point_cam[:2]) / d_safe
   p_norm2 = p.squared_norm()
   p_norm4 = p_norm2 * p_norm2
   r = 1 + k1*p_norm2 + k2*p_norm4
   pixel_projected = focal_length * r * p
   return pixel_projected - px_m
\end{minted}
In the code above, we define the node types \mintinline{python}{Cam} and \mintinline{python}{Point}, representing camera poses and landmark positions, as well as the observation type \mintinline{python}{Pixel}.
The residual function computes the reprojection error for a given camera and point, and we annotate the arguments to indicate which are tunable, which are constant, and how they should be accessed in memory.

From this factor definition alone, Caspar generates 43 CUDA kernels, a C++ solver that invokes them, a CMake file, and Python bindings so the solver can be called from both C++ and Python.
Since the residual formulation is symbolic, it is easy to test different weighting schemes or robust loss functions, but for comparison, we use the trivial reprojection error, consistent with other solvers.
Adding factor types such as IMU factors, GNSS pseudorange residuals, or model priors requires only defining additional residual functions, though this is not applicable to the \gls{bal} problem.

\subsection{Performance Comparison}

\subsubsection*{Speed}
As shown in Figure~\ref{fig:performance}, Caspar achieves a significant speed advantage over all competing solvers across the tested datasets.
This is particularly pronounced on the larger datasets, where Caspar is up to 20x faster than the next best solver at reaching certain accuracies.
Caspar is also surprisingly fast on the smaller datasets, significantly outperforming the CPU solvers.
Profiling indicates that the memory optimizations described in Sections~\ref{sec:symbolic_programming}--\ref{sec:symbolic_memory} are the primary contributors to Caspar's performance gains.
In an earlier version of Caspar, we used cuDSS from NVIDIA to solve the linearized problem.
This version was slower than the current one, but still faster than Ceres, indicating that both Caspar's faster linearization and residual evaluation, as well as its custom \gls{pcg} implementation, contribute to the speed.

\subsubsection*{Memory}
Figure~\ref{fig:performance} also indicates that the Caspar solver exhibits superior memory efficiency relative to competing solvers across the tested datasets.
The only exception is the Schur variation of Ceres, which is one of the slowest solvers.
This memory efficiency enables the solution of larger problem instances within the same hardware constraints.

\subsubsection*{Accuracy}
Although Caspar is not always the most precise solver, the final \gls{mse} values show that the accuracy is comparable to other solvers and has not been sacrificed for speed.
This small difference in accuracy is likely due to numerical errors arising from using float32 when aggregating large sets of values.
It should be noted that the values shown in Figure~\ref{fig:performance} are reported directly by the solvers, and that the true costs may differ slightly for solvers using float32, owing to numerical inaccuracies during summation.

\section{Conclusion}
In this paper, we presented Caspar, a new library for transforming symbolic expressions into optimized CUDA kernels and solving unconstrained nonlinear optimization problems on the GPU.
By bridging the gap between high-level mathematical formulation and low-level hardware execution, Caspar establishes performance-oriented symbolic programming as an expressive and effective alternative to manual kernel development in applicable domains.
Caspar allows researchers to symbolically formulate both individual kernels and nonlinear solvers from Python, and generates the infrastructure to run them from either Python or C++.

Representing symbolic operations as a \glsfirst{dabseg} enables functions with multiple outputs, a capability necessary for expressing hardware-mapped operations such as vectorized memory load instructions.
Building on this graph representation, Caspar applies targeted symbolic optimizations and performs expression reordering to minimize register pressure.
This symbolic compilation pipeline, incorporating memory optimizations through tailored accessors and vectorized memory layouts, leads to highly efficient kernels.
We demonstrated this by generating a nonlinear solver for bundle adjustment that is \textbf{5 to 20 times faster} than the best-performing GPU-accelerated alternative on the \gls{bal} dataset, while using less memory and achieving similar accuracy.

While these results clearly establish Caspar's end-to-end performance advantages on the \gls{bal} dataset, this work represents the first step in a broader research effort.
Moving forward, we intend to expand upon this foundation by exploring new application areas and conducting deeper component-based performance analysis.
Ultimately, our goal is to establish Caspar as the preferred solution for GPU-accelerated nonlinear optimization in robotics.

\section{Acknowledgements}
The conception and early developments of Caspar were funded by the Research Council of Norway through SFI AUTOSHIP (project number 309230).
The remaining developments were carried out during a paid PhD research internship at Skydio.
ChatGPT 4.1 and Claude Sonnet 4 were used to improve the language in this paper.

A special thanks to Hayk Martiros, Paul Gresia, and Edmund Brekke for reading through the paper and giving useful feedback.
Also, thanks to Caspar Wessel (\textdagger 1818) for his discoveries and for letting us use his name.
\printbibliography

\end{document}